\documentclass{article}

 \usepackage[preprint]{camprobe}

\usepackage[utf8]{inputenc} 
\usepackage[T1]{fontenc}    
\usepackage{hyperref}       
\usepackage{url}            
\usepackage{booktabs}       
\usepackage{amsfonts}       
\usepackage{nicefrac}       
\usepackage{microtype}      
\usepackage{xcolor}         
\usepackage{multirow}
\usepackage{array}
\usepackage{makecell}
\usepackage{pifont}
\usepackage{graphicx}
\usepackage{amsmath}
\usepackage{amssymb}
\usepackage{algorithm}
\usepackage{algpseudocode}
\usepackage{enumitem}
\usepackage[capitalize]{cleveref}
\newcommand{\p}[1]{\phantom{#1}}
\newcommand{\plusv}[1]{(+\textit{#1})}
\newcommand{\minusv}[1]{(-\textit{#1})}
\newcommand{\cmark}{\ding{51}}
\newcommand{\xmark}{\ding{55}}
\hypersetup{
    colorlinks=false, 
    linkcolor=blue, 
    citecolor=blue,   
    urlcolor=blue,    
    filecolor=magenta, 
}
\title{Probing into Camera Control of Video Models}

\author{%
  Chen Hou \\
  Visual Geometry Group\\
  University of Oxford\\
  \And
  Christian Rupprecht \\
  Visual Geometry Group\\
  University of Oxford \\
}

\begin{document}

\maketitle

\begin{abstract}
Video is a rich and scalable source of 3D/4D visual observations, and camera control is a key capability for video generation models to produce geometrically meaningful content. Existing approaches typically learn a mapping from camera motion to video using additional camera modules and paired data. However, such datasets are often limited in scale, diversity, and scene dynamics, which can bias the model toward a narrow output distribution and compromise the strong prior learned by the base model. These limitations motivate a different perspective on camera control.  In this paper, we show that camera control need not be modeled as an implicit mapping problem, but can instead be treated as a form of geometric guidance that induces displacements across frames. Specifically, we reformulate camera control into a set of displacement fields and apply them via differentiable resampling of latent features during denoising. Our simple approach achieves effective camera control with minimal degradation across diverse quality metrics compared to fine-tuned baselines. Since our method is applicable to most video diffusion models without training, it can also serve as a probe to study the camera control capabilities of base models. Using this probe, we identify universal biases shared by representative video models, as well as disparities in their responses to camera control. Finally, we benchmark their performance in multi-view generation, offering insights into their potential for 3D/4D tasks. \footnote{Webpage: {\hypersetup{hidelinks}\begingroup\color{magenta}\url{https://xrchitech.github.io/camprobe-page/}\endgroup}\\\hspace*{1.8em}Code: {\hypersetup{hidelinks}\begingroup\color{magenta}\url{https://github.com/xrchitech/CamProbe}\endgroup}}

\end{abstract}


\section{Introduction}
As the most common data modalities for capturing the real world, video naturally encodes both spatial structure and temporal dynamics, making it one of the richest sources of 3D and 4D visual signals. 
A key mechanism for accessing such stereoscopic structure lies in camera motion, which defines how a scene is observed over time. 
Enabling controllable camera motion is therefore crucial for bridging scalable 2D foundation models with 3D/4D understanding and generation, and is essential for applications such as view synthesis, downstream 3D reconstruction, and building the world model.

Recent approaches typically achieve camera control by fine-tuning video models on camera-pose-annotated datasets~\cite{bai2025recammaster, he2024cameractrl, yu2025trajectorycrafter}. 
The model is trained to learn a direct mapping from camera trajectories to generated videos. However, such a paradigm presents several fundamental limitations. 
First, the relationship between camera motion and generated content is implicitly encoded in the model and remains largely uninterpretable. 
Since only few geometric constraints are imposed, models may learn dataset-specific correlations rather than the underlying camera geometry.
Second, the available camera-annotated datasets are often limited in scale, diversity, and dynamics, as they are typically generated by rendering engines or Structure-from-Motion pipelines~\cite{bai2025recammaster, reizenstein2021co3d, zhou2018stereo}. 
Such datasets usually lack realistic motion patterns, long temporal dynamics, and complex scene variations, which restricts the range of camera motions and scenarios the model can generalize to. 
Third, fine-tuning on such specialized datasets tends to bias the model towards a narrower output distribution. 
This adaptation can override the rich and diverse priors learned during large-scale pretraining, leading to reduced visual diversity and degraded generation quality. Some training-free camera control methods~\cite{hou2024training} avoid such problems by manipulating latent feature layouts but rely heavily on non-differentiable operations, such as point cloud reconstruction and inpainting, thereby rendering end-to-end optimization intractable.

These limitations motivate a different perspective on camera control: since video diffusion models are already capable of generating realistic content, camera motion, and temporal dynamics, much of the complexity of video generation has been implicitly learned during pretraining. 
As a result, camera motion need not be modeled as a new generative process requiring training; it can instead be treated as a form of geometric guidance. Building on this insight, we reformulate camera control as a set of displacement fields integrated into the diffusion process. 
Given a camera trajectory, we first derive displacement fields that encode the spatial transformations induced by camera motion, then apply them via differentiable resampling of latent features during denoising. The displacement fields can either be heuristically defined or learned through end-to-end optimization. In this work, we show that depth-based geometry provides an effective means of constructing displacement fields. The approach enables effective camera control for video models with minimal degradation in quality across diverse evaluation metrics, even without training.

Beyond camera control, our formulation also serves as a probe for analyzing the camera control capabilities of video diffusion models. 
Effective evaluation of camera control in video generation remains underexplored. 
Current benchmarks are either designed for video understanding tasks with VLMs~\cite{lin2025camerabench} or rely on manually defined criteria for camera motion, which often fail to give correct assessments~\cite{zheng2025vbench2}. 
Moreover, in most cases, camera control is driven by text control. This setup not only entangles camera control with text-conditioned generation but, more importantly, does not faithfully reflect the model's camera control capability, as prompts often fail to induce the intended motion. 
However, our experiments suggest that these prompt-induced failures do not necessarily reflect an inability of the models, but instead arise from limitations of the guidance mechanism (Section~\ref{sec:exp-single-view}).
Using our probe, we draw several interesting conclusions, including shared biases across models in motion types and directions, as well as notable disparities in their response to camera control. 
We further evaluate their performance in generating multi-view geometry, providing insights into their potential use for downstream 3D/4D tasks.

In summary, our contributions are:
\begin{itemize}[leftmargin=1.5em, itemsep=0pt, topsep=-2pt]
\item We reformulate camera control as a set of displacement fields applied via differentiable resampling during denoising, enabling camera control without any training or additional modules.
\item We propose using this formulation as a \emph{probe} to study the camera control capability of video foundation models, thereby isolating model behavior from induction artifacts.
\item Using this probe, we identify several universal biases shared across representative video models, including underestimated camera control capability, asymmetric translation-rotation leakage, horizontal directional preference, as well as clear disparities in their sensitivity to camera control.
\item We introduce a multi-level evaluation protocol (2D, 2.5D, and 3D) for assessing multi-view consistency in video generation and benchmark five video models under orbital camera motions.
\end{itemize}




\section{Related Work}
\paragraph{Camera-controllable video generation.} 
Following the success of video models, camera-controllable video generation has attracted increasing attention~\cite{he2024cameractrl, hou2024training, bai2025recammaster}. Existing methods explore camera control from several directions. A primary line of research focuses on how to introduce camera signals into video models through various conditioning mechanisms, including adapter-based modules (e.g., LoRA~\cite{hu2022lora})~\cite{guo2023animatediff, zhao2024motiondirector}, camera-aware embeddings~\cite{he2024cameractrl, wang2024motionctrl, bahmani2024vd3d, bahmani2025ac3d}, and modifications to attention ~\cite{bai2025recammaster, hu2024motionmaster, kuang2024collaborative, xu2024cavia}. These approaches are typically trained via fine-tuning to learn a direct mapping from input trajectories to output videos. In addition, some works focus on dataset construction, curating or synthesizing videos with camera annotations to support supervised training~\cite{bai2024syncammaster, bai2025recammaster, reizenstein2021co3d, he2025cameractrl}. Another line of work incorporates explicit structures~\cite{yu2025trajectorycrafter, ren2025gen3c, xie2026lavr, feng2024i2vcontrol, you2024nvs, hou2024training} or geometric constraints~\cite{xu2024camco, kupyn2025epipolar} into the generation process, thereby providing inherent consistency in the generated visual signals.

\paragraph{Generative novel view synthesis.} 
Pretrained 2D models have been widely used for 3D/4D content generation. Existing approaches mainly follow several paradigms: fine-tuning feed-forward models for view synthesis~\cite{voleti2024sv3d, xie2024sv4d, bahmani2024vd3d, sun2024dimensionx}, optimizing explicit representations via score distillation sampling~\cite{poole2022dreamfusion, lin2023magic3d, liu2023zero, jiang2023consistent4d}, and generating dense multi-view images, which are then used for downstream 3D reconstruction~\cite{han2024vfusion3d, han2024flex3d, li2024vivid, wu2025cat4d}. These methods either rely on costly optimization or curated multi-view datasets with limited scale and diversity, making them less effective for complex scenes. Recent feed-forward reconstruction models have improved scalability by directly predicting scene geometry~\cite{wang2025vggt, lin2025depth}. However, modeling complex 4D dynamics remains challenging, as existing representations still struggle to match the fidelity and richness of real-world videos.

\paragraph{Benchmark for video camera control.} While several commonly used metrics exist 
for video quality assessment~\cite{unterthiner2018fvd, radford2021clipsim}, evaluation of camera control in video generation tasks is largely underexplored. Existing camera-video benchmarks fall into two categories. Some are designed for video understanding tasks, where camera motion is inferred using VLMs~\cite{lin2025camerabench}. Others rely on manually defined criteria to evaluate camera behavior~\cite{zheng2025vbench2}. For instance, VBench2.0~\cite{zheng2025vbench2} evaluates multi-view consistency by prompting camera motion through text descriptions and measuring the trajectories of tracked points using heuristic rules. In practice, such manually designed metrics are insufficient to capture more complex motion behaviors and often fail to provide reliable assessments.

\section{Method}
\begin{figure}[t]
    \centering
    \includegraphics[width=1\linewidth]{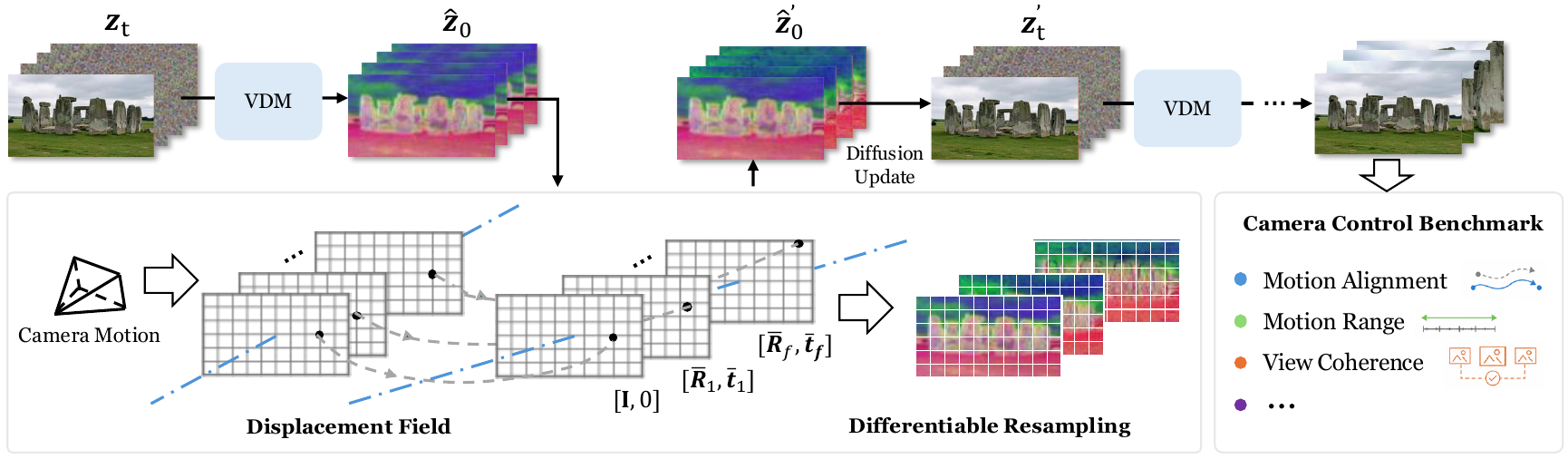}
    \caption{Our proposed method. Camera control is formulated as a set of displacement fields applied to update the denoising process via differentiable resampling. This also serves as a probe to study the camera control capabilities of video foundation models.}
    \label{fig:method}
\end{figure}

In this section, we present CamProbe, a method for training-free camera control, with its formulation and its use as a probe for camera control analysis. An overview is shown in Figure~\ref{fig:method}.

\subsection{Problem formulation}
Given an input image $\mathbf{x}_0$ and a camera trajectory $(T_f)_{f=1}^F, T_f \in \mathrm{SE}(3)$ of $F$ frames, the goal is to generate a video sequence $(\mathbf{x}_f)_{f=1}^F$ that follows the desired camera motion. Most existing methods learn a mapping $\left(x_0, (T_f)_{f=1}^F\right)\xrightarrow{\mathcal{M}}(\mathbf{x}_f)_{f=1}^F$ using additional camera modules $\mathcal{M}$ and paired data. Instead, we aim to find a set of displacement fields $\mathcal{F}$ for each camera motion $(T_f)_{f=1}^F\rightarrow\mathcal{F}$, then merge the displacement field into the generation process $\mathcal{F}\rightarrow(\mathbf{x}_f)_{f=1}^F$. The two steps are independent, and the displacement field can be learned or heuristically designed. In this paper, we show that simple depth-based warping already serves as an effective form of mapping $(T_f)_{f=1}^F\rightarrow\mathcal{F}$ and leave the learned variants for future work. 

\subsection{Camera control by displacement field}
\paragraph{From camera motion to displacement field.}
\label{sec:method-1}
Given a camera trajectory $\mathbf{T}_f=(\mathbf{R}_f,\mathbf{t}_f)$, with rotation $\mathbf{R}_f \in \mathrm{SO(3)}$, and translation $\mathbf{t}_f \in \mathbb{R}^3$, we construct a displacement field $\mathcal{F}_f$ for each frame that maps the pixel coordinates under a canonical view to their positions under the target camera motion. Since the absolute camera pose of the input image is unknown, we operate on relative transformations with respect to the first frame: $\bar{\mathbf{T}}_f = \mathbf{T}_f \mathbf{T}_1^{-1}$.
Considering a normalized coordinate $\mathbf{u}=[u, v, 1]^\top$ in the latent space $\mathbf{z}_f \in \mathbb{R}^{H \times W \times C}$, we first lift it to a pseudo-3D point in the canonical coordinate system using off-the-shelf depth estimator:
\begin{equation}
    \mathbf{p}_f(\mathbf{u}) = D_f(\mathbf{u}) \cdot \mathbf{K}^{-1} \mathbf{u},
\end{equation}
where $D_f(\mathbf{u})$ denotes the depth map, and $\mathbf{K}$ is a basic camera intrinsic matrix, which is assumed to have a unit focal length and zero principal point in the normalized coordinate space. In practice, depth estimation may introduce scale ambiguity across frames and videos; however, we empirically observe that such ambiguity has a negligible impact on performance (Section~\ref{ablation}). We then apply the camera transformation and project the point back to the image plane:
\begin{equation}
    \mathbf{p}_f'(\mathbf{u})=\bar{\mathbf{R}}_f\mathbf{p}_f(\mathbf{u}) + \bar{\mathbf{t}}_f, \quad
    \mathbf{u}'=\mathbf{\Pi}(\mathbf{K}\mathbf{p}_f'(\mathbf{u})),
\end{equation}
where $\mathbf{\Pi}$ denotes the perspective projection operator.
The displacement field is then defined as:
\begin{equation}
    \mathcal{F}_f := \mathbf{u}'-\mathbf{u}.
\end{equation}
This formulation yields a dense warping grid for each coordinate driven by input camera motion. 
More importantly, the proposed formulation does not restrict the displacement field to be analytically defined. Since warping is implemented via differentiable sampling (e.g., backward grid sampling), it can be further parameterized and optimized with supervision from input video and camera motion.
\paragraph{Diffusion update.}
Given displacement fields $\mathcal{F}$, we integrate them into the iterative denoising process. At each diffusion timestep $t$, an estimation of the clean sample $\mathbf{\hat{z}}_0$ can be derived from the current noisy latent $\mathbf{z}_t$ and the model output. For each frame $f$ of $\mathbf{\hat{z}}_0$, we apply the displacement field independently in a classifier-free guidance~\cite{ho2022classifier} form:
\begin{equation}
    \hat{\mathbf{z}}_{0,f}' = \hat{\mathbf{z}}_{0,f} + \omega\left(\mathcal{F}_f \circ \hat{\mathbf{z}}_{0,f}-\hat{\mathbf{z}}_{0,f}\right) .
\end{equation}
Here, $\circ$ denotes a differentiable resampling operator. Note that we apply each displacement field to its respective frame rather than propagating from a single reference. This ensures that the intrinsic dynamics and content synthesized by the pretrained model are preserved, while only their spatial positions are refined to produce an effective camera motion. Empirically, we find that a scale of $\omega=1$ is most effective, since the guide is in pure geometric form.

The modified estimate $\mathbf{\hat{z}}_{0,f}'$ is not directly used for the next sampling step, but to update the current noisy latent by reimposing diffusion noise. The updated latent under noise level $\sigma_t$ can be derived as:
\begin{equation}
    \mathbf{z}_t'=(1-\sigma_t) \hat{\mathbf{z}}_0'+\sigma_t\boldsymbol{\epsilon}.
\end{equation}
\begin{minipage}[t]{0.48\textwidth}
For velocity prediction~\cite{saharia2022photorealistic}, we update the velocity jointly since it contains both the signal and noise. Taking flow matching~\cite{lipman2022flow} as an example:
\begin{equation}
\begin{aligned}
\mathbf{v}_t'=\epsilon&-\hat{\mathbf{z}}_0', \\
\mathbf{z}_t'=\hat{\mathbf{z}}_0'&+\sigma_t\mathbf{v}_t'.
\end{aligned}
\end{equation}
The noise $\boldsymbol{\epsilon}$ can either be recovered from unmodified latent $\mathbf{z}_t$ or  sampled from a Gaussian distribution $\boldsymbol{\epsilon}\sim\mathcal{N}(0, \mathbf{I})$. In most cases, we find that modifying $v$ alone without re-calculating $\mathbf{z}_t'$ is already sufficient for effective control (Section~\ref{ablation}), thus this step can be omitted.
This procedure is applied iteratively during the early stage of the denoising process, typically for $t\in[T, 0.8T]$, which works consistently across different video diffusion models. The complete algorithm is summarized in Algorithm~\ref{alg:main}.
\end{minipage}
\hfill
\begin{minipage}[t]{0.48\textwidth}
\vspace{-15pt}
\begin{algorithm}[H]
\caption{Diffusion update (v-prediction)}
\label{alg:main}
\begin{algorithmic}[1]
\State \textbf{Input:} \parbox[t]{.8\linewidth}{Reference image $\mathbf{x}_0$, model $f_\theta$, \\ trajectory $(\mathbf{T}_f)_{f=1}^F$, sampler $\mathcal{S}$}
\State Sample $\mathbf{z}_T \sim \mathcal{N}(0, \mathbf{I})$

\For{$t = T, \dots, 1$}
    \State $\hat{\mathbf{z}}_0, \mathbf{v}_t \leftarrow f_\theta(\mathbf{z}_t, t, \mathbf{x}_0)$
    \If{$t \in [T, 0.8T]$}
        \For{$f = 1, \dots, F$}
            \State Compute $\mathcal{F}_f$
            \State $\hat{\mathbf{z}}_{0,f}' = \mathcal{F}_f \circ \hat{\mathbf{z}}_{0,f} $
        \EndFor
        \State Sample $\boldsymbol{\epsilon} \sim \mathcal{N}(0, \mathbf{I})$ 
        \State $\mathbf{v}_t' = \boldsymbol{\epsilon} - \hat{\mathbf{z}}_0'$
        \State $\mathbf{z}_t \leftarrow \hat{\mathbf{z}}_0' + \sigma_t \mathbf{v}_t'$ \Comment{Optional}
    \EndIf
    \State $\mathbf{z}_{t-1} \leftarrow \mathcal{S}(\mathbf{z}_t, \mathbf{v}_t', t)$
\EndFor
\end{algorithmic}
\end{algorithm}
\end{minipage}

\subsection{Interpretation as a probe}
Since our method enables camera control without introducing additional modules or fine-tuning, it can serve as an analytical tool for probing the camera control capability of video diffusion models. Specifically, the displacement fields impose geometric control while leaving content and dynamics generation entirely to the pretrained model. This minimum intervention allows us to isolate how models respond to controlled camera motion and assess their ability to adapt to viewpoint changes. Using this probe, we study the behavior of representative video models from two perspectives:
\begin{itemize}[left=0pt]
    \item \textbf{Single-view motions}, including pan, tilt, truck, pedestal, and dolly/zoom. By applying displacement fields at the same scale, we evaluate the extent to which each model can handle camera motions of different types and directions. We observe several universal biases shared across models, as well as clear disparities in their responses to different camera motions. 
    \item \textbf{Multi-view motions}, e.g., orbital trajectories. We design a set of metrics, ranging from image-level to geometry-level, to assess alignment and consistency in multi-view generation, and benchmark five video foundation models under these settings.
\end{itemize}

Together, this provides a comprehensive characterization of camera control capabilities in video diffusion models, allowing us to assess their controllability under commonly used camera motions and their potential for downstream 3D/4D generation and reconstruction tasks.

\paragraph{Does the probe measure the model or our method?} Our method can be viewed as a simple, standardized geometric intervention that probes model \emph{responses} to camera control. The displacement field is fixed and applied identically across all models without adaptation, while content, temporal dynamics, and novel-view synthesis are still governed by the pretrained video model. Therefore, the different behaviors observed under the same probe primarily reflect differences in how models respond to camera motion, as evidenced by the substantial variation across models in our experiments.

\section{Experiments}

In this section, we evaluate our camera control method and compare it with state-of-the-art approaches. We then use it as a probe to study the camera control behavior of several video models.

\subsection{Video camera control}
\label{sec:exp-compare}
\paragraph{State-of-the-art comparison.} We first compare our method with state-of-the-art camera control approaches in Table~\ref{tab:part1-i2v} (detailed configurations in Table~\ref{supp:table-checkbox}). As most video-based models do not support video-to-video (v2v) generation, we ensure comparable content across models as follows: we first use an image-to-video (i2v) base model to generate a reference video for v2v methods, and then use its first frame as input for i2v-based camera control. We consider a balanced set of camera motions, including truck right, pan left, tilt up, pedestal down, zoom in, zoom out, arc left, and arc right. Each motion is paired with 93 prompts from the ``overall consistency'' category in VBench~\cite{huang2024vbench}. Camera control performance is computed using DepthAnything3~\cite{lin2025depth}, under normalized coordinates in the range of $[-1, 1]$ and averaged over multiple temporal window sizes (1,4,8,12). We use $\mathit{Sim}(3)$ alignment as each method produces different scales of motion. Compared with fine-tuned approaches, our method does not rely on large-scale fine-tuning or explicit 3D representations, yet achieves competitive visual quality with a minor drop in camera control accuracy.

\paragraph{Base model comparison.} We further evaluate the effect of camera control by comparing performance before and after its application. To ensure comparable content, we use Wan2.1-T2V (the base model of ReCamMaster~\cite{bai2025recammaster}) to generate reference videos, and take their first frames as input to our base model (HunyuanVideo-I2V). Evaluation is conducted on 10 trajectories from the official ReCamMaster repository\footnote{https://github.com/KlingAIResearch/ReCamMaster} covering both basic rotations and orbital motions. For each trajectory, we use the same prompts as in the previous experiment. Results are reported in Table~\ref{tab:part1-t2v}. When comparing against baselines, we report raw RPE without alignment, as $\mathit{Sim}(3)$ tends to favor small but steady motions and may thus unfairly improve baseline performance. While both methods improve camera control accuracy over base models, our method better preserves video quality, with less degradation across most metrics. This suggests that effective camera control does not necessarily require retraining the video model. In contrast, fine-tuning-based methods may compromise the pretrained generative prior and lead to reduced quality. Displacement fields provide a natural solution that preserves the base model's original performance. Additional results are provided in Appendix~\ref{supp: full-metric-base}. 
\begin{figure}[t]
    \centering
    \includegraphics[width=1\linewidth]{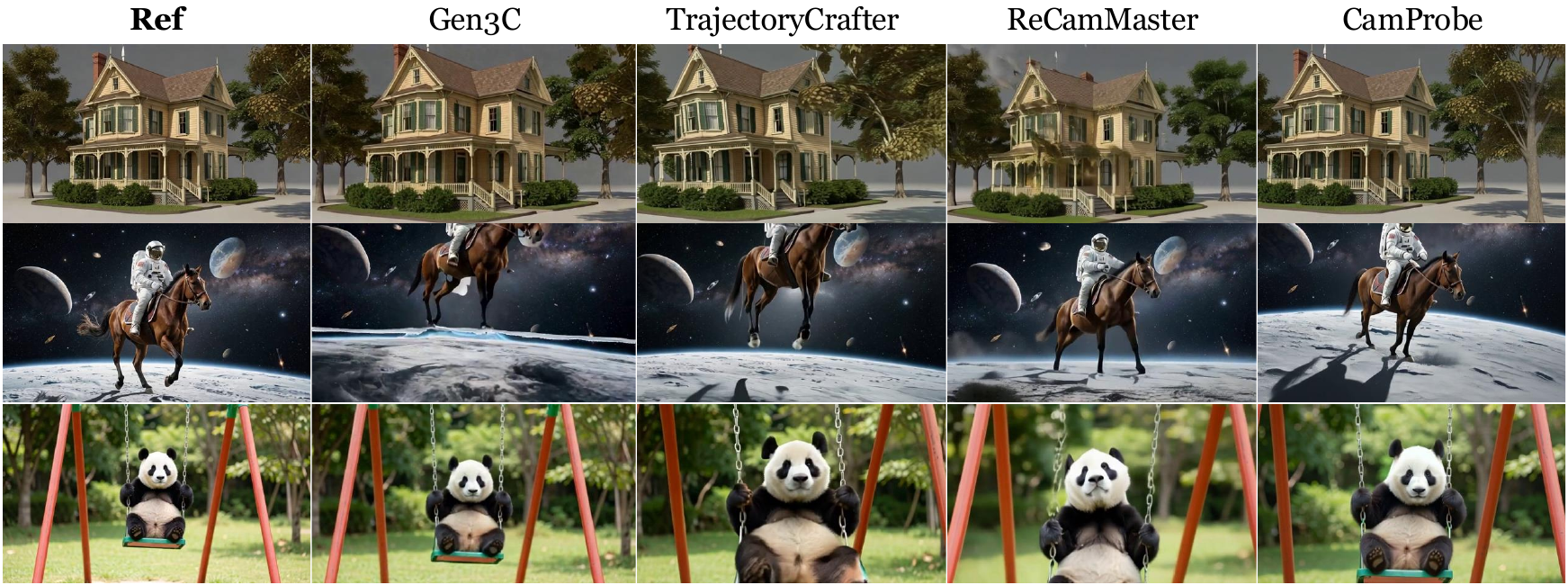}
    \caption{Qualitative comparison. Motions from top to bottom: truck right, pedestal down, zoom in. Fine-tuned methods can lead to deformations, irrational content, or unsuccessful motions.}
    \label{fig: compairson}
\end{figure}
\begin{table*}[t]
\centering
\caption{State-of-the-art comparison. Full metrics are reported in Appendix~\ref{supp: full-metric-sota}.}
\label{tab:part1-i2v}
\small
\begin{tabular}{lcccc|cc}
\toprule
\multicolumn{1}{c}{\multirow{2}{*}{\raisebox{-3ex}{Method}}}
& \multicolumn{4}{c|}{\textbf{Video Quality}}
& \multicolumn{2}{c}{\textbf{Camera Control}} \\
\cmidrule(lr){2-5}
\cmidrule(lr){6-7}
& \makecell[c]{Dynamic \\ Degree $\uparrow$}
& \makecell[c]{Imaging \\ Quality $\uparrow$}
& \makecell[c]{Motion \\ Smoothness $\uparrow$}
& \makecell[c]{Background \\ Consistency $\uparrow$}
& \makecell[c]{RPE-T $\downarrow$}
& \makecell[c]{RPE-R $\downarrow$} \\
\midrule

Gen3C~\cite{ren2025gen3c}
& 51.08
& 66.09
& 99.21
& 96.25
& 0.032
& 3.12 \\

TrajectoryCrafter~\cite{yu2025trajectorycrafter}
& 54.30
& 68.40
& 99.20
& 96.14
& 0.021
& \textbf{2.73} \\

ReCamMaster~\cite{bai2025recammaster}
& 46.37
& 67.33
& \textbf{99.24}
& 95.65
& \textbf{0.018}
& 3.23 \\

CamProbe (Ours)
& \textbf{55.24}
& \textbf{68.83}
& 99.08
& \textbf{96.28}
& 0.039
& 2.95 \\

\bottomrule
\end{tabular}
\end{table*}
\subsection{Camera control capabilities of foundation models}
\label{sec:exp-probe}
We use our method as a probe to evaluate the camera control capabilities of several state-of-the-art video generation models. We study their behavior under two categories of motion: single-view motions, which mainly preserve the original scene content, and multi-view motions, which require synthesizing novel viewpoints. For single-view motions, we consider a set of in-plane camera motions with balanced types and directions, including pan right, truck left, tilt up, and zoom out. We select 30 prompts from the prompt set used in the previous section, covering diverse content such as outdoor scenes, animals, humans, close-up objects, and synthetic content (Appendix~\ref{supp: selected-prompts}). For multi-view motions, we evaluate clockwise and counterclockwise orbital motions around the y-axis using the 97 prompts from the ``multi-view consistency’’ category of VBench2.0~\cite{zheng2025vbench2}. To ensure fair evaluation, we remove the camera motion descriptions and rely solely on our method to induce camera motion. All models are evaluated under identical settings for resolution, denoising steps, warping steps, and warping scale. First-frame images are generated using FLUX 2.0~\cite{flux-2-2025} to avoid potential bias from models that may have been trained on overlapping data. See Appendix~\ref{supp: exp-setting} for more details.
\begin{table*}[t]
\centering
\caption{Quantitative comparison with base models. Full metrics are reported in Appendix~\ref{supp: full-metric-base}.}
\label{tab:part1-t2v}
\small
\setlength{\tabcolsep}{4pt}
\begin{tabular}{lcccc|cc}
\toprule
\multicolumn{1}{c}{\multirow{2}{*}{\raisebox{-3ex}{Method}}}
& \multicolumn{4}{c|}{\textbf{Video Quality}}
& \multicolumn{2}{c}{\textbf{Camera Control}} \\
\cmidrule(lr){2-5}
\cmidrule(lr){6-7}
& \makecell[c]{Dynamic \\ Degree $\uparrow$}
& \makecell[c]{Imaging \\ Quality $\uparrow$}
& \makecell[c]{Motion \\ Smoothness $\uparrow$}
& \makecell[c]{Background \\ Consistency $\uparrow$}
& \makecell[c]{RPE-T $\downarrow$}
& \makecell[c]{RPE-R $\downarrow$} \\
\midrule

\textbf{Wan2.1-T2V}
& 32.26\p{(+\textit{00.00})}
& 69.42\p{(+\textit{0.00})}
& 98.83\p{(+\textit{0.00})}
& 96.72\p{(+\textit{0.00})}
& 0.090
& 2.228 \\

+ ReCamMaster
& 58.74\plusv{26.48}
& 67.62\minusv{1.80}
& 98.92\textbf{\plusv{0.10}}
& 94.52\minusv{2.20}
& \textbf{0.077}
& \textbf{1.371} \\


\midrule

\textbf{HunyuanVideo-I2V}
& 32.26\p{(+\textit{00.00})}
& 68.87\p{(+\textit{0.00})}
& 99.24\p{(+\textit{0.00})}
& 95.64\p{(+\textit{0.00})}
& 0.111
& 2.240 \\

+ CamProbe (Ours)
& 59.41\textbf{\plusv{27.15}}
& 68.43\textbf{\minusv{0.43}}
& 99.01\minusv{0.23}
& 95.17\textbf{\minusv{0.46}}
& 0.104
& 1.706 \\

\bottomrule
\end{tabular}
\end{table*}
\subsubsection{Single-view motion}
\label{sec:exp-single-view}
\begin{figure}[t]
    \centering
    \includegraphics[width=0.85\linewidth]{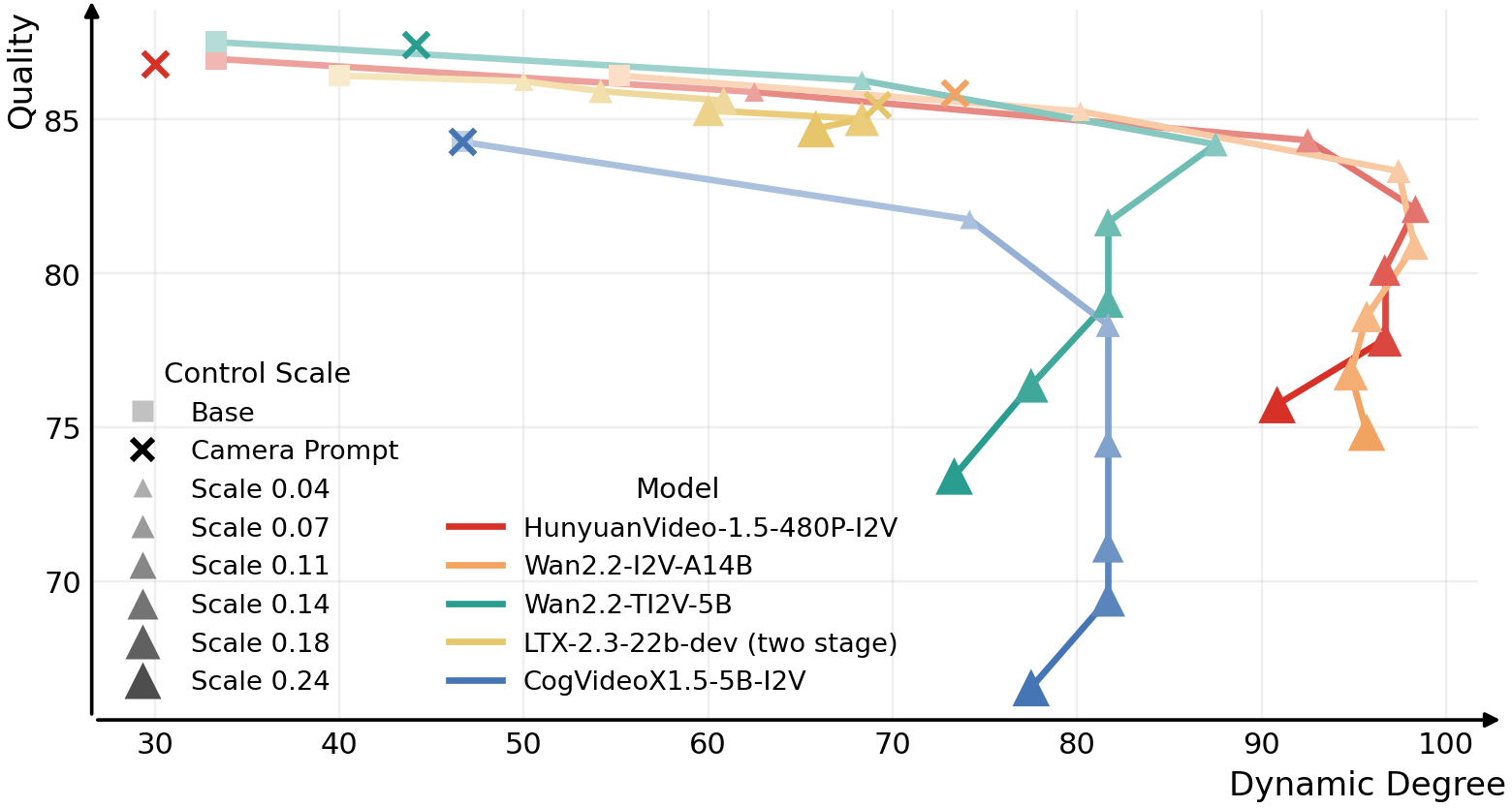}
    \caption{Trade-off between dynamic degree and quality across video models under varying camera control scales. When the motion scale exceeds a certain threshold, the dynamics counterintuitively decrease due to a shift in motion mode, where smooth camera motion is replaced by abrupt transitions.}
    \label{fig: dynamic-quality trade-off}
\end{figure}

\paragraph{Camera motion bias.} As we progressively increase the camera control scale, we examine how video quality changes accordingly. Here, video quality is defined as the average of all quality metrics except dynamic degree (Appendix~\ref{supp: full-metric-sota}). The control scale specifies the relative strength of the camera motion used to construct the displacement field. Since the warping operates in normalized coordinate $[-1,1]$, the scale is not tied to a physical camera distance, but only controls how strongly camera motion is induced. This allows us to plot camera motion-quality curves for different models, as shown in Figure~\ref{fig: dynamic-quality trade-off}. Note that base models already exhibit inherent dynamics (e.g., object motion), so the gap between camera-controlled results and base outputs reflects the effect of camera motion. From these curves, we observe several common trends across models:
\begin{itemize}[left=0pt]
    \item \textbf{Capability bias.} First, we find that existing evaluations may underestimate the camera control capability of video diffusion models. Prior work~\cite{zheng2025vbench2} often reports poor camera control performance under prompt-based settings (e.g., low accuracies in VBench 2.0), where camera control is entangled with text-conditioned generation, which often fails to induce the intended motion. However, our results in Figure~\ref{fig: dynamic-quality trade-off} show that, although prompt-based control is only moderately effective (marked with $\times$), most models can generate substantially stronger camera motion with only minor quality degradation under our method. This suggests that the limitation lies not in the models themselves, but in how camera motion is induced during generation.
    \item \textbf{Translation is easier than rotation.} Second, we observe that translational motions are consistently easier for models to handle than rotational ones. 
    As shown in Table~\ref{tab:motion_leakage}, motion leakage is asymmetric: under rotational commands, much of the predicted motion is translational, while the reverse is smaller. This asymmetry likely arises because translation is approximately linear displacement, whereas rotation induces spatially varying transformations that are harder to generate. Note that the SfM-based estimator tends to underestimate translation. Therefore, the measured translation leakage is likely conservative and the true asymmetry may be stronger. The values are averaged across all tested models. Detailed metric definitions and per-model results are provided in Appendix~\ref{supp:leakage}.
    \item \textbf{Horizontal is easier than vertical.} Third, we observe directional biases in camera motion. Specifically, movements in horizontal directions are handled more effectively than vertical ones, in both dynamics and quality preservation (Appendix~\ref{supp:xy}). This aligns with the bias in real-world videos, where camera motion more frequently occurs along horizontal directions than vertical ones.
    \item \textbf{Motion mode shift.} When the magnitude of camera control exceeds a certain threshold, the resulting motion dynamics counterintuitively decrease. This behavior is typically caused by a mode shift in generation, where smooth camera motion is replaced by abrupt transitions, effectively switching between different viewpoints or scene content. See Appendix~\ref{supp:video-results} for examples.
    \item \textbf{Universal trade-off.} Finally, we observe systematic trade-offs between motion strength and visual quality. While larger control scales induce stronger camera motion dynamics, they may also degrade visual quality, with different models exhibiting varying sensitivity to this trade-off.
\end{itemize}


\paragraph{Controllability across different models.} Among all tested models, HunyuanVideo-1.5 and Wan2.2-I2V-A14B achieve the best balance between camera motion strength and visual quality, 
with Wan2.2-I2V-A14B exhibiting a later motion mode shift. 
Wan2.2-TI2V-5B shows comparable visual quality but degrades earlier in dynamics as camera control strength increases. LTX-2.3-22b maintains the highest quality but shows limited camera control capability, while CogVideoX1.5-5B suffers the most severe degradation in quality under strong control. Detailed results are reported in Appendix~\ref{supp:probe-results} and \ref{supp:video-results}.

\subsubsection{Multi-view motion}
\label{sec:exp-multi-view}
\paragraph{Multi-view geometry.} We evaluate multi-view capabilities from two aspects: (1) whether a model can follow novel viewpoints as intended, and (2) whether the generated views are geometrically consistent. For viewpoint control, we measure camera motion alignment using Relative Pose Error, as well as the cosine similarity between the translation and rotation vectors and their corresponding ground-truth axes. For geometric consistency, we evaluate at three levels:

\begin{enumerate}[itemsep=0pt, topsep=-3pt,left=0pt]
\item \textbf{Image level} (2D): we report the RMSE of the epipolar error in pixel space, using the Fundamental matrix estimated from ground-truth camera poses and SIFT correspondences~\cite{lowe2004distinctive}.
\item \textbf{Depth-aware level} (2.5D): we estimate depth using DepthAnything 3~\cite{lin2025depth} and compute the warping RMSE in normalized RGB space by warping the input frame to subsequent frames.
\item \textbf{Gaussian splatting level} (3D): we reconstruct scenes using Gaussian Splatting~\cite{kerbl20233d} and report reprojection RMSE in normalized RGB space on 20 randomly sampled prompts.
\end{enumerate}

\begin{figure}[t]
    \centering
    \includegraphics[width=1\linewidth]{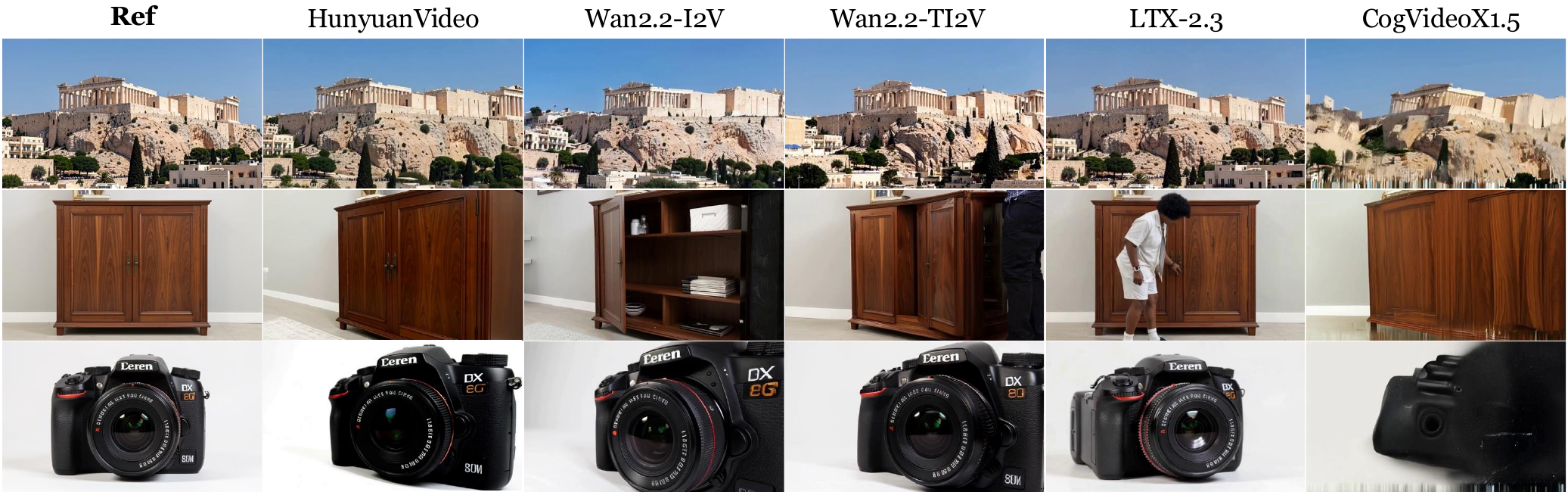}
    \caption{Visualization of probing the multi-view  capabilities of base models. Motions: arc right.}
    \label{fig: multi-view-probe}
\end{figure}

In addition, we randomly sample 100 objects from Objaverse-XL~\cite{deitke2023objaverse}, render them as videos using the same orbital trajectory, and then evaluate the same metrics to establish reference upper and lower bounds. As shown in Table~\ref{tab:part2-3d}, LTX-2.3 and HunyuanVideo-1.5 achieve relatively strong performance in multi-view consistency. However, the consistency metrics tend to reward static content and penalize dynamic effects, including both camera motion and object motion. Therefore, consistency should be considered jointly with orbital alignment. This behavior is also reflected in the visual results (Figure~\ref{fig: multi-view-probe}), where LTX-2.3 tends to produce high-quality videos but with relatively limited motion. In contrast, HunyuanVideo-1.5 and Wan2.2-I2V-A14B achieve a better balance between geometric consistency and motion alignment, demonstrating stronger performance in novel view synthesis.

\subsection{Method Analysis}
\label{ablation}

\begin{table}[t]
\small
\centering
\begin{minipage}{0.35\linewidth}
\centering
\caption{Motion leakage.}
\begin{tabular}{lc}
\toprule
\textbf{Type} & \textbf{Leakage} $\downarrow$ \\
\midrule
Rot. $\rightarrow$ Trans. & 6.04 \\
Trans. $\rightarrow$ Rot. & \textbf{1.08} \\
\bottomrule
\end{tabular}
\label{tab:motion_leakage}
\end{minipage}
\hspace{-2em}
\begin{minipage}{0.65\linewidth}
\centering
\caption{Horizontal and vertical bias.}
\begin{tabular}{lcc}
\toprule
\textbf{Motion} & \textbf{Dynamic Inc. (\%)} $\uparrow$ & \textbf{Quality Dec. (\%)} $\downarrow$ \\
\midrule
Horizontal & +\textbf{110.98} & -\textbf{1.68} \\
Vertical & +42.22 & -2.05 \\
\bottomrule
\end{tabular}
\label{tab:direction_bias}
\end{minipage}
\end{table}

\paragraph{Single-reference latent replacement.} When the displacement field is applied only to the input image and the warped reference signal directly replaces $\hat{\textbf{z}}_0$ at a target step, our formulation degenerates to a setting similar to CamTrol~\cite{hou2024training}. However, there are several critical differences between the two methods. First, CamTrol constructs warped frames from a single input image, causing subsequent frames to largely propagate the first frame's content and suppress the intrinsic dynamics of the generated signal. More importantly, CamTrol relies on point cloud reconstruction and inpainting pipelines to guide layout changes, making direct end-to-end optimization intractable. In contrast, our method directly revises the latent by differentiable resampling throughout the denoising process, allowing the displacement field to be naturally parameterized and optimized with supervision from paired data. Experiments show that, under comparable visual quality, our method achieves stronger dynamics than CamTrol. Detailed results are provided in Appendix~\ref{supp:camtrol}.

\paragraph{Diffusion update.}
We compare three diffusion update strategies: (1) updating both $\hat{\textbf{z}}_0$ and $\textbf{v}_t$, followed by re-sampling back to $\textbf{z}_t'$; (2) updating $\hat{\textbf{z}}_0$ and $\textbf{v}_t$ while keeping $\textbf{z}_t$ unchanged; and (3) updating only $\textbf{v}_t$: $\textbf{v}_{t}'=\mathcal{F}_f \circ \textbf{v}_t$. We find that the strategy (2) already yields competitive performance. This property is desirable as it simplifies future end-to-end optimization by avoiding backpropagation through the denoising network. However, keeping $\mathbf{z}_t$ unchanged can introduce spatial inconsistency between the original signals in $\mathbf{z}_t$ and the warped signals in $\mathbf{v}_t'$, leading to occasional ghosting artifacts (see Appendix~\ref{supp:diffusion-update}). Directly warping the entire velocity signal in strategy (3), including both $\hat{\mathbf{z}}_0$ and the noise component, severely disrupts the latent distribution and results in unstable generation. 

\paragraph{Depth normalization.} We evaluate different depth normalization strategies described in Section~\ref{sec:method-1}, including raw depth, sequence-level (across all frames), and per-frame normalization. Depth maps are estimated using MiDaS~\cite{ranftl2020towards} from the decoded $\hat{\textbf{z}}_0'$, and we also consider a constant-depth setting where all values are set to 1. As normalization has a limited impact in single-view scenarios, we focus on the multi-view setup (Section~\ref{sec:exp-multi-view}). Quantitatively, differences across settings are minor, and even constant depth achieves comparable results. However, this comes at the cost that the warping degrades to a homography, leading to severe dragging effects and reduced image quality. For stability, we use per-frame normalization in this paper. Nevertheless, the effectiveness of the constant-depth setting suggests that even very coarse guidance can still steer video models to produce camera-motion effects. Detailed comparison results are provided in Appendix~\ref{supp:depth-norm}.

\begin{table*}[t]
\centering
\small
\caption{Multi-view geometry comparison.}
\label{tab:part2-3d}
\begin{tabular}{
l
>{\centering\arraybackslash}m{1.5cm} 
>{\centering\arraybackslash}m{1.2cm} |
>{\centering\arraybackslash}m{1.2cm}
>{\centering\arraybackslash}m{1.2cm}
>{\centering\arraybackslash}m{1.2cm}
}
\toprule
\multicolumn{1}{c}{\multirow{2.5}{*}{Method}}
& \multicolumn{2}{c|}{\textbf{Orbit Alignment}}
& \multicolumn{3}{c}{\textbf{Multi-view Consistency}} \\
\cmidrule(lr){2-3} \cmidrule(lr){4-6} 
& RPE-T/-R $\downarrow$ & Axis $\uparrow$ & 2D $\downarrow$ & 2.5D $\downarrow$ & 3D $\downarrow$ \\ 
\midrule

HunyuanVideo-1.5-480P-I2V~\cite{hunyuanvideo2025}{}
& \underline{0.054}/\textbf{4.69} & \underline{0.62}
& \underline{34.20} & \textbf{0.2565} & 0.2123 \\

Wan2.2-I2V-A14B~\cite{wan2025wan}
& \textbf{0.046}/\underline{4.90} & \textbf{0.64}
& 50.02 & \underline{0.2689} & 0.2204 \\

Wan2.2-TI2V-5B~\cite{wan2025wan}
& 0.072/5.04 & 0.55
& 39.90 & 0.2777 & \underline{0.2108} \\

LTX-2.3-22b-dev (two stage)~\cite{hacohen2026ltx}
& 0.097/5.24 & 0.33
& \textbf{27.72} & 0.2801 & \textbf{0.1840} \\

CogVideoX1.5-5B-I2V~\cite{yang2024cogvideox}
& 0.079/5.22 & 0.53
& 70.38 & 0.2807 & 0.2531 \\

\textbf{\textcolor{gray}{Objaverse-XL~\cite{deitke2023objaverse} (lower/upperbound)}}
& \textbf{\textcolor{gray}{0.015/1.51}} & \textbf{\textcolor{gray}{0.88}}
& \textbf{\textcolor{gray}{27.61}} & \textbf{\textcolor{gray}{0.0998}} & \textbf{\textcolor{gray}{0.0905}} \\

\bottomrule
\end{tabular}
\end{table*}

\section{Conclusion and Limitations}
\label{sec:conclusion-limitation}
In this paper, we show that camera control in video generation need not be treated as an implicit learning problem, but is instead more naturally solved through displacement field-guided generation. Such displacement fields themselves constitute a sufficiently strong signal for camera control that even a heuristically defined one can serve as an effective method without any fine-tuning. This suggests that learnable formulations may bring further improvements in both controllability and precision. Using it as a probe, we systematically study the camera control capabilities of video foundation models and their potential for 3D/4D generation.

A limitation of our method is that the displacement field is predefined rather than learned. While this enables effective control for simple motions and even handles more complex movements, it becomes less reliable in scenes with complex spatial relationships (see Appendix~\ref{supp:video-results} for examples). Second, the current formulation relies heavily on the base model's generative prior. As shown in Section~\ref{ablation}, using a constant depth still yields comparable results, indicating that the geometric guidance is coarse rather than precise. These observations indicate that further improvements may point to a learnable displacement field or more effective diffusion updates.

\newpage
\section*{Acknowledgement}
This project was supported by ERC starting grant `Volute' (No. 101222037).


{
\small
\bibliographystyle{plain}
\bibliography{ref}


}

\newpage
\appendix
\section*{\LARGE Appendix}
\section{Model configurations}
\label{supp-model-config}
Table~\ref{supp:table-checkbox} summarizes the model configurations. Unlike prior works that rely on large-scale fine-tuning on curated datasets, our method is training-free and directly applied to pretrained video models.
\begin{table*}[h]
\caption{Model configurations.}
\label{supp:table-checkbox}
\setlength{\tabcolsep}{1pt}
\centering
\begin{tabular}{lcccc}
\hline
Method & \textbf{Fine-tune} & \textbf{Base Model} & \textbf{Dataset} & \textbf{Size} \\
\hline
GEN3C~\cite{ren2025gen3c}              & \cmark & Cosmos-Predict1-7B-Video2World~\cite{agarwal2025cosmos} & \makecell[c]{RE10K~\cite{zhou2018stereo}\\DL3DV~\cite{ling2024dl3dv}\\WOD~\cite{sun2020scalability}\\Kubric4D~\cite{greff2022kubric}} & \textasciitilde200k \\
\midrule
TrajectoryCrafter~\cite{yu2025trajectorycrafter}  & \cmark & CogVideoX-Fun-V1.1-5b-InP~\cite{yang2024cogvideox} & \makecell[c]{OpenVid-1M~\cite{nan2024openvid}\\DL3DV~\cite{ling2024dl3dv}\\RealEstate10K~\cite{zhou2018stereo}} & 180K \\
\midrule
ReCamMaster~\cite{bai2025recammaster}        & \cmark & Wan2.1-T2V-1.3B~\cite{wan2025wan} & \makecell[c]{MultiCamVideo~\cite{bai2025recammaster}} & 136K \\
\midrule
CamProbe           & \xmark & \makecell[c]{HunyuanVideo-1.5-480P-I2V~\cite{hunyuanvideo2025}\\(Any)} & \xmark & \xmark \\
\hline
\end{tabular}
\end{table*}
\section{Experiment settings}
\label{supp: exp-setting}
Most experiment settings are described in Section~\ref{sec:exp-compare} and Section~\ref{sec:exp-probe}. For all experiments using our method, including both comparison and probing settings, we adopt a unified configuration with a resolution of $832 \times 480$, 49 frames, 25 denoising steps, and 5 diffusion update steps. We use DepthAnything 3~\cite{lin2025depth} for estimating the camera poses of generated videos, and all reported RPE metrics are averaged over temporal windows of sizes 1, 4, 8, and 12 for sequences of 49 frames. For video generation, we use imageio\footnote{https://github.com/imageio/imageio.} to write frames for all models and set the frame rate to 16 FPS, as VBench is sensitive to video encoding strategies. 

Our method does not require training or introduce additional trained modules, and therefore incurs no additional training cost. To evaluate the inference overhead, we compare the runtime and memory consumption of the base model with and without our method. We use HunyuanVideo-1.5-480P-I2V~\cite{hunyuanvideo2025} for testing, and report the average per-video runtime over 10 generated videos. The update interval is set to $[T, 0.8T]$. As shown in Table~\ref{supp:time-gpu}, the additional GPU memory introduced by displacement-field warping is negligible compared with the video diffusion backbone itself. The main overhead comes from runtime rather than GPU memory, largely due to depth estimation and latent resampling. When using the diffusion update strategy without resampling $\mathbf{z}_t'$ (see Section~\ref{ablation} and Appendix~\ref{supp:diffusion-update}), the runtime drops noticeably; similarly, replacing decoded depth estimation with constant depth (Section~\ref{ablation}, Appendix~\ref{supp:depth-norm}) also reduces the runtime significantly. When both strategies are applied together, the additional runtime overhead is almost entirely eliminated.

\begin{table}[h]
\centering
\label{supp:time-gpu}
\caption{Inference overhead comparison.}
\label{tab:runtime-memory}
\begin{tabular}{lcc}
\toprule
Method & Runtime (s) & GPU Peak Memory (GB) \\
\midrule
\textbf{Base Model} & \textbf{48.69} & \textbf{50.65} \\
+ ours & 80.65 & 50.75 \\
+ ours (w/o $\hat{\textbf{z}}_t$ resampling) & 65.70 & 50.74 \\
+ ours (const. depth) & 63.49 & 50.66 \\
+ ours (w/o $\hat{\textbf{z}}_t$ resampling + const. depth) & 48.69 & 50.66 \\
\bottomrule
\end{tabular}
\end{table}

\section{Full metric state-of-the-art comparison}
\label{supp: full-metric-sota}

\begin{table*}[h]
\centering
\caption{Comparison with state-of-the-art camera control method.}
\label{supp:full-sota}
\small
\setlength{\tabcolsep}{5pt}

\begin{tabular}{lcccccc}
\toprule
Method
& \makecell[c]{Dynamic \\ Degree $\uparrow$}
& \makecell[c]{Imaging \\ Quality $\uparrow$}
& \makecell[c]{Motion \\ Smoothness $\uparrow$}
& \makecell[c]{Background \\ Consistency $\uparrow$}
& \makecell[c]{Subject \\ Consistency $\uparrow$}
& \makecell[c]{Aesthetic \\ Quality $\uparrow$} \\
\midrule

Gen3C~\cite{ren2025gen3c}
& 51.08
& 66.09
& 99.21
& 96.25
& 94.86
& 63.45 \\

TrajectoryCrafter~\cite{yu2025trajectorycrafter}
& 54.30
& 68.40
& 99.20
& 96.14
& 96.26
& 64.45 \\

ReCamMaster~\cite{bai2025recammaster}
& 46.37
& 67.33
& \textbf{99.24}
& 95.65
& 96.24
& 63.52 \\

CamProbe
& \textbf{55.24}
& \textbf{68.83}
& 99.08
& \textbf{96.28}
& \textbf{96.26}
& \textbf{66.49} \\

\bottomrule
\end{tabular}
\end{table*}


We provide the full metric comparison with state-of-the-art camera control methods in Table~\ref{supp:full-sota}. As shown, our method achieves the best performance on most visual quality metrics, including dynamic degree, imaging quality, and consistency. 

\section{Full metric comparison with base models}
\label{supp: full-metric-base}
We report comparison results with baselines in full metric from VBench (for custom prompt video) in Table~\ref{tab:supp-full-base}. Since our method can be seamlessly incorporated into any video diffusion model, we also include comparisons where we use the same base model as ReCamMaster, i.e., Wan2.1-T2V. Results show that it can achieve comparable quality with the fine-tuned method.
\begin{table*}[htbp]
\centering
\caption{Full metric comparison with baselines.}
\label{tab:supp-full-base}
\scriptsize
\setlength{\tabcolsep}{6pt}

\begin{tabular}{lcccccc}
\toprule
Method
& \makecell[c]{Dynamic \\ Degree $\uparrow$}
& \makecell[c]{Imaging \\ Quality $\uparrow$}
& \makecell[c]{Motion \\ Smoothness $\uparrow$}
& \makecell[c]{Background \\ Consistency $\uparrow$}
& \makecell[c]{Subject \\ Consistency $\uparrow$}
& \makecell[c]{Aesthetic \\ Quality $\uparrow$} \\
\midrule

\textbf{Wan2.1-T2V-1.3B}~\cite{wan2025wan}
& 32.26\p{ (+\textit{0.00})}
& 69.42\p{ (+\textit{0.00})}
& 98.83\p{ (+\textit{0.00})}
& 96.72\p{ (+\textit{0.00})}
& 97.66\p{ (+\textit{0.00})}
& 62.09\p{ (+\textit{0.00})} \\

+ ReCamMaster~\cite{bai2025recammaster}
& 58.74 \plusv{26.48}
& 67.62 \minusv{1.80}
& 98.92 \textbf{\plusv{0.10}}
& 94.52 \minusv{2.20}
& 95.63 \minusv{2.03}
& 58.96 \minusv{3.14} \\

+ CamProbe
& 82.26 \textbf{\plusv{50.00}}
& 67.89 \minusv{1.53}
& 98.86 \minusv{0.04}
& 94.53 \minusv{2.19}
& 94.36 \minusv{3.31}
& 60.67 \textbf{\plusv{1.42}} \\

\midrule

\textbf{HunyuanVideo-I2V}~\cite{hunyuanvideo2025}
& 32.26\p{ (+\textit{0.00})}
& 68.87\p{ (+\textit{0.00})}
& 99.24\p{ (+\textit{0.00})}
& 95.64\p{ (+\textit{0.00})}
& 96.49\p{ (+\textit{0.00})}
& 59.97\p{ (+\textit{0.00})} \\

+ CamProbe
& 59.41 \plusv{27.15}
& 68.43 \textbf{\minusv{0.43}}
& 99.01 \minusv{0.23}
& 95.17 \textbf{\minusv{0.46}}
& 95.41 \textbf{\minusv{1.07}}
& 61.17 \plusv{1.20} \\

\bottomrule
\end{tabular}
\end{table*}










\section{Quantitative probing results}
\label{supp:probe-results}
We report the quantitative results of the dynamics-quality trade-offs across all tested models. Specifically, we compare (1) the base dynamics of videos generated by the original models, (2) the results obtained using camera prompts, and (3) the maximum achievable camera motion under our probing method, together with the corresponding visual quality at that point.

The camera prompts used for the four evaluation motions are:
\begin{enumerate}[label={}, itemsep=0pt, topsep=-3pt, leftmargin=15pt]
\item Pan right:~~~The camera smoothly pans to the right.
\item Tilt up:~~~~~~~The camera smoothly tilts upward.
\item Truck left:~~The camera moves left with a smooth lateral translation.
\item Zoom out:~~The camera smoothly zooms out.
\end{enumerate}

\begin{table}[h]
\centering
\caption{Dynamics-quality trade-offs.}
\label{tab:dynamic_summary}
\begin{tabular}{
l
>{\centering\arraybackslash}m{1.2cm}
>{\centering\arraybackslash}m{1.5cm}
>{\centering\arraybackslash}m{1.2cm}
>{\centering\arraybackslash}m{2cm}
}
\toprule
Model
& Base Dyn. ($\uparrow$)
& Prompt Dyn. ($\uparrow$)
& Max Dyn. ($\uparrow$)
& Qual. at Max Dyn. ($\uparrow$) \\
\midrule
HunyuanVideo-1.5-480P-I2V~\cite{hunyuanvideo2025} & 33.33 & 30.00 & \textbf{98.33} & 82.09 \\
Wan2.2-I2V-A14B~\cite{wan2025wan} & \textbf{55.17} & \textbf{73.33} & 98.28 & 80.89 \\
Wan2.2-TI2V-5B~\cite{wan2025wan} & 33.33 & 44.17 & 87.50 & \textbf{84.19} \\
LTX-2.3-22b-dev~\cite{hacohen2026ltx} & 40.00 & 69.17 & 68.33 & 85.02 \\
CogVideoX1.5-5B-I2V~\cite{yang2024cogvideox} & 46.67 & 46.67 & 81.67 & 78.32 \\
\bottomrule
\end{tabular}
\end{table}

\section{Translation-Rotation leakage}
\label{supp:leakage}
In this section, we provide the detailed definition of the translation–rotation leakage metric used in Section~\ref{sec:exp-single-view}.
Given a sequence of predicted camera poses $(\mathbf{R}_f, \mathbf{t}_f)_{f=1}^F$, we first compute the relative motion between adjacent frames:
\begin{equation}
    \Delta \mathbf{R}_f = \mathbf{R}_{f+1} \mathbf{R}_f^\top, \quad
\Delta \mathbf{t}_f = \mathbf{t}_{f+1} - \mathbf{t}_f.
\end{equation}
We then calculate the rotation vector converted from relative rotational pose $\boldsymbol{\omega}_f = \log(\Delta \mathbf{R}_f)$. To account for the mismatch between rotational and translational units (i.e., radians and spatial distance), we use a scaling factor $\lambda$ that converts rotation into a translation-equivalent magnitude. Specifically, we compute the average magnitude of ground-truth rotation over rotation-only motions $R_{\text{ref}}$ (e.g., pan and tilt), and the average magnitude of ground-truth translation over translation-only motions $D_{\text{ref}}$ (e.g., truck and zoom). Their ratio is used as a scaling factor to align the two motion types:
\begin{equation}
\lambda = \frac{D_{\text{ref}}}{R_{\text{ref}}}.
\end{equation}

Under single-motion settings, we define leakage as the proportion of unintended motion relative to the ground-truth motion magnitude. For pure translational sequences (e.g., zoom or truck), where ground-truth rotation is negligible, we measure rotational leakage as:
\begin{equation}
L_{\text{trans.}\rightarrow\text{rot.}}^{\text{gt}} =
\frac{
\lambda \sum_f \|\boldsymbol{\omega}_f^{pred}\|
}{
\sum_f \|\Delta \mathbf{t}_f^{gt}\| + \epsilon
}.
\end{equation}

For pure rotational sequences, where ground-truth translation is negligible, we measure translational leakage as:
\begin{equation}
L_{\text{rot.}\rightarrow\text{trans.}}^{\text{gt}} =
\frac{
\sum_f \|\Delta \mathbf{t}_f^{pred}\|
}{
\lambda \sum_f \|\boldsymbol{\omega}_f^{gt}\| + \epsilon
}.
\end{equation}

We use these metrics to analyze the relative relationship between $rot.\rightarrow trans.$ and $trans.\rightarrow rot.$ leakage under a consistent motion command scale. However, when comparing individual models, this formulation may be biased by the overall motion magnitude of the predictions. In particular, models that produce smaller motions tend to exhibit smaller leakage values when normalized by ground-truth motion (LTX-2.3 in Table~\ref{tab:supp_leakage}). To address this, we additionally define a normalized leakage ratio with respect to the predicted motion magnitude:
\begin{equation}
L_{\text{trans.}\rightarrow\text{rot.}}^{\text{pred}} =
\frac{
\lambda \sum_f \|\boldsymbol{\omega}_f^{pred}\|
}{
\sum_f \|\Delta \mathbf{t}_f^{pred}\| + \epsilon
},\quad 
L_{\text{rot.}\rightarrow\text{trans.}}^{\text{pred}} =
\frac{
\sum_f \|\Delta \mathbf{t}_f^{pred}\|
}{
\lambda \sum_f \|\boldsymbol{\omega}_f^{pred}\| + \epsilon
}.
\end{equation}
Here we set $\epsilon$ as 1e-9. Both $L_{\text{trans.}\rightarrow\text{rot.}}$ and $L_{\text{rot.}\rightarrow\text{trans.}}$ are dimensionless and quantify the amount of unintended motion relative to the intended motion under a unified scale. The scaling factor $\lambda$ is fixed across all models and is computed solely from ground-truth motion statistics, ensuring a fair and consistent comparison. We apply these metrics under single-motion settings and report per-model results in Table~\ref{tab:supp_leakage}. The results show that translation–rotation leakage is a consistent and systematic bias across all evaluated models. The results in the main paper (Table~\ref{tab:motion_leakage}) are computed under the ground-truth (GT) normalization, as our goal is to compare the relative strength of leakage across the two directions, rather than to compare individual models.

Note that all models are evaluated under the same ground-truth motion scale, so the metrics allow for fair relative comparison across models, as well as between the two leakage directions ($Rot.\rightarrow Trans.$ and $Trans.\rightarrow Rot.$). However, since rotation and translation are measured in different units, the absolute values of these metrics do not have a direct physical meaning, and cannot be used to define an exact equivalence between rotation and translation. Instead, they should be interpreted as relative indicators that reflect how strongly a model exhibits cross-motion leakage under a consistent evaluation setting. Here, our analysis focuses on relative comparisons rather than absolute magnitudes.

\begin{table}[h]
\centering
\caption{Motion leakage under translation and rotation movements.}
\label{tab:supp_leakage}
\footnotesize
\begin{tabular}{
l
>{\centering\arraybackslash}m{1.7cm}
>{\centering\arraybackslash}m{1.7cm}
|
>{\centering\arraybackslash}m{1.7cm}
>{\centering\arraybackslash}m{1.7cm}
}
\toprule
\multirow{2}{*}{Model}
& \multicolumn{2}{c|}{\textbf{Pred.}\quad $\downarrow$}
& \multicolumn{2}{c}{\textbf{GT.}\quad $\downarrow$} \\
\cmidrule(lr){2-3}
\cmidrule(lr){4-5}
& Rot. $\rightarrow$ Trans.
& Trans. $\rightarrow$ Rot.
& Rot. $\rightarrow$ Trans.
& Trans. $\rightarrow$ Rot.\\
\midrule

HunyuanVideo-1.5-480P-I2V~\cite{hunyuanvideo2025}
& 3.49 & \textbf{0.41} & \underline{4.75} & \underline{0.73} \\

Wan2.2-I2V-A14B~\cite{wan2025wan}
& \textbf{2.92} & 0.44 & 7.33 & 1.62 \\

Wan2.2-TI2V-5B~\cite{wan2025wan}
& \underline{3.40} & 0.51 & 5.93 & 1.18 \\

LTX-2.3-22b-dev~\cite{hacohen2026ltx}
& 9.23 & 0.48 & \textbf{4.31} & \textbf{0.55} \\

CogVideoX1.5-5B-I2V~\cite{yang2024cogvideox}
& 3.41 & \underline{0.42} & 7.86 & 1.32 \\

\bottomrule
\end{tabular}
\end{table}

\section{Horizontal-Vertical preference}
\label{supp:xy}
To calculate the preference of model's camera motion for horizontal versus vertical camera motion, we use two rotational motions of the same angle with different directions: tilt up and pan right. We adopt the base model output (without camera control) as reference $d_{\text{base}},q_{\text{base}}$, and measure the dynamics increase and quality decrease when applying the same scale of camera motion at different directions:
\begin{equation}
        D\_inc=\frac{d_{\text{camera}} - d_{\text{base}}}{d_{\text{base}}}, \quad Q \_ dec=    \frac{q_{\text{base}} - q_{\text{camera}}}{q_{\text{base}}}.
\end{equation}
Here, quality is defined as the average of five metrics: subject consistency, background consistency, motion smoothness, aesthetic quality, and imaging quality (excluding dynamic degree).

\begin{table*}[h]
\centering
\small
\caption{Directional bias and relative changes.}
\label{tab:supp_direction_bias}
\begin{tabular}{
l
>{\centering\arraybackslash}m{2cm}
>{\centering\arraybackslash}m{3cm}
>{\centering\arraybackslash}m{3cm}
}
\toprule
\multirow{2}{*}{Method}
& \textbf{Base}
& \textbf{Pan right}
& \textbf{Tilt up} \\
\cmidrule(lr){2-2} \cmidrule(lr){3-3} \cmidrule(lr){4-4}
& Dyn / Qual
& Dyn (↑\%) / Qual (↓\%)
& Dyn (↑\%) / Qual (↓\%) \\
\midrule

HunyuanVideo-1.5-480P-I2V~\cite{hunyuanvideo2025} 
& 33.33 / 86.97 
& 93.33 (↑180.00\%) / 85.61 (↓1.56\%) 
& 53.33 (↑60.00\%) / 85.56 (↓1.62\%) \\

Wan2.2-I2V-A14B~\cite{wan2025wan}
& 55.17 / 86.43 
& 93.10 (↑68.75\%) / 85.01 (↓1.64\%) 
& 65.52 (↑18.75\%) / 84.88 (↓1.79\%) \\

Wan2.2-TI2V-5B~\cite{wan2025wan}
& 33.33 / 87.52 
& 93.33 (↑180.00\%) / 85.97 (↓1.76\%) 
& 60.00 (↑80.00\%) / 85.77 (↓2.00\%) \\

LTX-2.3-22b-dev~\cite{hacohen2026ltx}
& 40.00 / 86.42 
& 53.33 (↑33.33\%) / 86.20 (↓0.26\%) 
& 46.67 (↑16.67\%) / 86.30 (↓0.14\%) \\

CogVideoX1.5-5B-I2V~\cite{yang2024cogvideox}
& 46.67 / 84.27 
& 90.00 (↑92.86\%) / 81.56 (↓3.21\%) 
& 63.33 (↑35.71\%) / 80.31 (↓4.70\%) \\
\bottomrule
\end{tabular}
\end{table*}
As shown in Table~\ref{tab:supp_direction_bias}, nearly all the models tested show the same extent of bias, where the horizontal direction movement cause larger dynamics and smaller quality damage compared with the vertical direction motion of the same scale.



    
    

\section{Comparison with CamTrol}
\label{supp:camtrol}
Although both are training-free methods for camera control in video generation, our approach differs from CamTrol~\cite{hou2024training} in several key aspects:

\begin{enumerate}[itemsep=0pt, topsep=0pt, leftmargin=15pt]
\item CamTrol constructs warped frames from a single input image, such that subsequent frames are largely propagated from the first frame. As a result, the $\hat{\textbf{z}}_0$ signal in $\textbf{z}_t$ can become overly static, limiting the intrinsic dynamics synthesized by the pretrained model. Instead, our method applies each displacement field independently to its corresponding frame. This preserves the original dynamics and content generated by the pretrained model, while refining only their spatial positions to induce effective camera motion.
\item More importantly, CamTrol relies on explicit point cloud reconstruction and inpainting pipelines to guide latent layout change. While such a formulation demonstrates that camera control can be induced through layout manipulation, the control signal itself is analytically constructed and tightly coupled with the external rendering pipeline, making direct end-to-end optimization intractable. In contrast, our method directly applies differentiable resampling to revise the latent throughout denoising. As a result, the displacement field is not restricted to analytically defined trajectories, but can instead be parameterized and optimized under supervised learning setting.
\end{enumerate}

As shown in Table~\ref{supp:tab-camtrol}, when achieving comparable visual quality to CamTrol, our method produces stronger camera-motion dynamics. Experiments are conducted on HunyuanVideo-1.5-480P-I2V with four camera motions, including pan right, tilt up, truck left, and zoom out.

\begin{table}[h]
\centering
\caption{Comparison with CamTrol.}
\label{supp:tab-camtrol}
\scriptsize
\begin{tabular}{lcccccc|cc}
\toprule
\multicolumn{1}{c}{\multirow{2}{*}{\raisebox{-3ex}{Method}}}
& \multicolumn{6}{c|}{\textbf{Video Quality}\quad $\uparrow$}
& \multicolumn{2}{c}{\textbf{Camera Control}\quad $\downarrow$} \\
\cmidrule(lr){2-7}
\cmidrule(lr){8-9}

& \makecell[c]{Dynamic \\ Degree }
& \makecell[c]{Aesthetic \\ Quality}
& \makecell[c]{Imaging \\ Quality}
& \makecell[c]{Motion \\ Smoothness}
& \makecell[c]{Background \\ Consistency}
& \makecell[c]{Subject \\ Consistency }
& \makecell[c]{RPE-T}
& \makecell[c]{RPE-R} \\

\midrule

CamTrol~\cite{hou2024training}
& 35.00
& 66.05
& \textbf{73.80}
& 98.62
& 95.43
& 95.65
& \textbf{0.904}
& 0.016 \\

CamProbe
& \textbf{62.50}
& \textbf{66.74}
& 72.27
& \textbf{98.76}
& \textbf{95.87}
& \textbf{95.79}
& 0.920
& \textbf{0.010} \\

\bottomrule
\end{tabular}
\end{table}

\section{Diffusion update}
\label{supp:diffusion-update}
We compare diffusion update strategies: 

\begin{enumerate}[label=(\arabic*), itemsep=1pt, topsep=0pt, leftmargin=25pt]
\item updating both $\hat{\textbf{z}}_0$ and $\textbf{v}_t$, followed by re-sampling back to $\textbf{z}_t'$;
\item updating $\hat{\textbf{z}}_0$ and $\textbf{v}_t$ while keeping $\textbf{z}_t$ unchanged;
\item updating only $\textbf{v}_t$: $\textbf{v}_{t}'=\mathcal{F}_f \circ \textbf{v}_t$.
\end{enumerate}

\begin{figure}[h]
    \centering
    \includegraphics[width=1\linewidth]{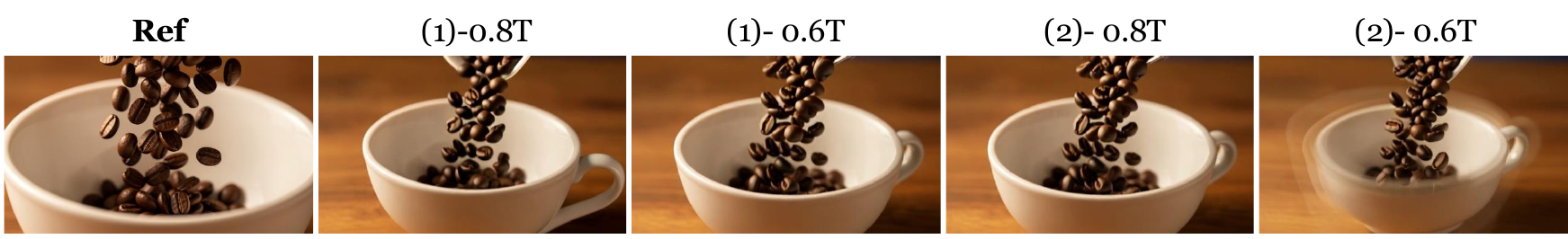}
    \caption{Comparison of update strategies under different denoising steps. Motions: zoom out.}
    \label{fig: supp-update}
\end{figure}

As shown in Figure~\ref{fig: supp-update}, strategy (2) already yields competitive performance compared with strategy (1). However, since the unchanged $\mathbf{z}_t$ still contains the original unmoved signal, sampling with the updated $\textbf{v}_{t}'$ can introduce spatial misalignment between two signals, occasionally leading to ghosting artifacts. This effect becomes more severe as the update interval increases (e.g., (2)-0.6T in Figure~\ref{supp:diffusion-update} which refers to diffusion update in$[T, 0.6T]$). Directly warping the entire velocity signal in strategy (3), including both $\hat{\mathbf{z}}_0$ and the noise component, severely disrupts the latent distribution and results in unstable generation. 

Despite this limitation, strategy (2) remains attractive because it avoids modifying $\mathbf{z}_t$, allowing optimization to be performed directly on the displacement field without back-propagating through the denoising network. This significantly simplifies future end-to-end optimization and reduces both memory and computational overhead. Moreover, strategy (2) also brings faster inference speed (Section~\ref{supp: exp-setting}).

\section{Depth normalization}
\label{supp:depth-norm}
We report the quantitative results of different depth normalization methods in Table~\ref{supp:tab-depth-norm}. In the raw setting, we directly use the predicted depth without normalization, which makes the warping sensitive to scale variations. The sequence-level strategy uses a shared scale across all frames, computed from the median depth of the first frame, while the per-frame strategy normalizes each frame independently using its own median depth. For the constant setting (1), we simply set all depth values to 1, and no depth estimation is needed. The base model used in this experiment is HunyuanVideo-1.5-I2V. Although the constant setting performs surprisingly well in both quality metrics and camera control accuracy, the warping collapse to a homography, which can lead to severe dragging effects and poorer image quality. For more stable results, we use per-frame setting in this paper.

\begin{table*}[h]
\label{supp:tab-depth-norm}
\centering
\caption{Effects of depth normalization.}
\scriptsize
\begin{tabular}{lcccccc|cc}
\toprule
\multicolumn{1}{c}{\multirow{2}{*}{\raisebox{-3ex}{Method}}}
& \multicolumn{6}{c|}{\textbf{Video Quality}}
& \multicolumn{2}{c}{\textbf{Camera Control}} \\
\cmidrule(lr){2-7}
\cmidrule(lr){8-9}

& \makecell[c]{Dynamic \\ Degree $\uparrow$}
& \makecell[c]{Imaging \\ Quality $\uparrow$}
& \makecell[c]{Motion \\ Smoothness $\uparrow$}
& \makecell[c]{Background \\ Consistency $\uparrow$}
& \makecell[c]{Subject \\ Consistency $\uparrow$}
& \makecell[c]{Aesthetic \\ Quality $\uparrow$}
& \makecell[c]{RPE-T $\downarrow$}
& \makecell[c]{RPE-R $\downarrow$} \\

\midrule

raw
& 38.66
& 71.78
& \textbf{99.33}
& 96.48
& 97.65
& 58.19
& 0.1005
& 3.0885 \\

1
& \textbf{54.30}
& 69.03
& 99.17
& \textbf{96.58}
& 96.51
& \textbf{66.59}
& \textbf{0.0927}
& \textbf{2.9994} \\

sequence
& 35.05
& 71.76
& 99.32
& \textbf{96.58}
& \textbf{97.67}
& 58.07
& 0.0930
& 3.0905 \\

perframe
& 37.63
& \textbf{71.82}
& \textbf{99.33}
& 96.54
& \textbf{97.67}
& 58.18
& 0.0974
& 3.1207 \\

\bottomrule
\end{tabular} 
\end{table*}

\begin{figure}[h]
    \centering
    \includegraphics[width=1\linewidth]{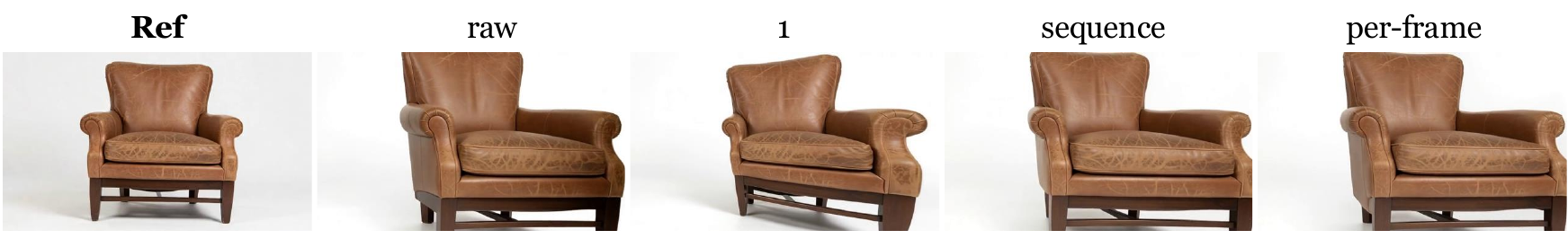}
    \caption{Comparison of depth norm. Constant depth setting can lead to severe dragging effects.}
    \label{fig: supp-norm}
\end{figure}


\section{Video results}
\label{supp:video-results}
Additional video results are available via: 

{\hypersetup{hidelinks}\begingroup\color{magenta}\url{https://xrchitech.github.io/camprobe-page/}\endgroup}.

The webpage includes videos of multi-view generation, basic motions, complex trajectories, probing results on base models, comparisons with state-of-the-art camera control methods, examples of motion mode shift, and examples of failure cases. 


\section{Codes}
\label{supp:code}
The code is provided via:

{\hypersetup{hidelinks}\begingroup\color{magenta}\url{https://github.com/xrchitech/CamProbe}\endgroup}. 

Our implementation is built upon the HunyuanVideo-1.5 inference framework~\cite{hunyuanvideo2025}. We extend its image-to-video generation by incorporating our displacement field and diffusion update in the generating process. The current code support four basic motions: pan right, tilt up, truck left, and zoom out, but can also take custom camera coordinates as input. Apart from the camera control mechanism, the implementation retains all original components of HunyuanVideo-1.5, including the original network structure and optional inference optimizations.


\section{Selected prompts for probing single-view motions}
\label{supp: selected-prompts}
In Section~\ref{sec:exp-probe}, we selected 30 prompts for each of the four single-view motions for evaluate the base models' capabilities. These prompts are drawn from the VBench "overall consistency" prompt set, which contains 97 prompts in total and is also used for comparisons with other camera control methods. We ensure coverage of diverse prompt categories to enable a fair and comprehensive evaluation across models. The selected prompts and their categories are as follows:

\paragraph{Outdoor scenes.}
\begin{enumerate}[label={}, itemsep=0pt, topsep=-3pt, leftmargin=15pt]
\item Yellow flowers swing in the wind.
\item Pacific coast, carmel by the sea ocean and waves.
\item Campfire at night in a snowy forest with starry sky in the background. 
\item A steam train moving on a mountainside. 
\item A drone flying over a snowy forest. 
\item The bund Shanghai, vibrant color. 
\item A modern art museum, with colorful paintings. 
\end{enumerate}

\paragraph{Human.}
\begin{enumerate}[label={}, itemsep=0pt, topsep=-3pt, leftmargin=15pt]
\item Vincent van Gogh painting in a room.
\item An oil painting of a couple in formal evening wear caught in heavy rain with umbrellas.
\item Gwen Stacy reading a book.
\item A close-up of an artist painting on a canvas with a brush.
\item A morning makeup routine.
\item An astronaut feeding ducks on a sunny afternoon, with reflections on the water.
\end{enumerate}

\paragraph{Animals.}
\begin{enumerate}[label={}, itemsep=0pt, topsep=-3pt, leftmargin=15pt]
\item A cat wearing sunglasses by a pool.
\item A teddy bear washing dishes.
\item A happy Corgi playing in a park at sunset.
\item A cat eating food from a bowl.
\item A turtle swimming in the ocean.
\item A jellyfish floating through the ocean with bioluminescent tentacles.
\item A panda drinking coffee in a café in Paris.
\end{enumerate}

\paragraph{Close-up objects.}
\begin{enumerate}[label={}, itemsep=0pt, topsep=-3pt, leftmargin=15pt]
\item An ashtray full of cigarette butts on a table, with smoke flowing against a black background, close-up.
\item Macro slow-motion close-up of roasted coffee beans falling into an empty bowl.
\item An ice cream melting on a table.
\end{enumerate}

\paragraph{Synthetic CG content.}
\begin{enumerate}[label={}, itemsep=0pt, topsep=-3pt, leftmargin=15pt]
\item A 3D model of a Victorian house from the 1800s.
\item A boat sailing along the Seine River with the Eiffel Tower in the background, in the style of Vincent van Gogh.
\item A coastal beach in spring with waves lapping on the sand, in the style of Hokusai (Ukiyo-e).
\item A coastal beach in spring with waves lapping on the sand, in the style of Vincent van Gogh.
\item A robot dancing in Times Square.
\item A confused panda in a calculus class.
\item A hyper-realistic spaceship landing on Mars.
\end{enumerate}

We use the augmented version of these prompts as in Vbench~\footnote{\url{https://github.com/Vchitect/VBench/blob/master/prompts/augmented_prompts/gpt_enhanced_prompts/prompts_per_dimension_longer/overall_consistency_longer.txt}}.

\section{Licenses}
\label{licenses}
\begin{itemize}[label=\textbullet, leftmargin=25pt, itemsep=0pt]
\item \makebox[5.8cm][l]{HunyuanVideo-1.5-480P-I2V:} Tencent Hunyuan Community License Agreement
\item \makebox[5.8cm][l]{Wan2.2-TI2V-5B/Wan2.2-I2V-A14B:} Apache 2.0 License
\item \makebox[5.8cm][l]{LTX-2.3-22b-dev:} LTX-2 Community License Agreement
\item \makebox[5.8cm][l]{CogVideoX1.5-5B-I2V:} Apache 2.0 License
\item \makebox[5.8cm][l]{DepthAnything 3:} Apache 2.0 License
\item \makebox[5.8cm][l]{MiDaS:} MIT License
\item \makebox[5.8cm][l]{FLUX 2.0:} Apache 2.0 License
\item \makebox[5.8cm][l]{GEN3C:} Apache 2.0 License
\item \makebox[5.8cm][l]{TrajectoryCrafter:} Custom non-commercial license
\item \makebox[5.8cm][l]{ReCamMaster:} MIT License
\end{itemize}



\end{document}